\documentclass[journal]{IEEEtran}

\usepackage{cite}
\usepackage{amsmath,amssymb,amsfonts}
\usepackage{booktabs}
\usepackage[caption=false,font=footnotesize]{subfig}
\usepackage{graphicx}
\usepackage{textcomp}
\usepackage{xcolor}
\usepackage{float}
\usepackage[section]{placeins}
\usepackage{enumitem}
\usepackage[utf8]{inputenc}
\usepackage{url}
\usepackage{xurl}
\usepackage[breaklinks,colorlinks]{hyperref}
\graphicspath{{./}}
\title{FERA: A Pose-Based Framework for Rule-Grounded Multimedia Decision Support with a Foil Fencing Case Study}

\author{%
Ziwen~Chen\\
Marc Garneau Collegiate Institute (TDSB), Toronto, Canada\\
\texttt{ziwen08@gmail.com}\\[0.75em]
Zhong~Wang\\
Beijing Institute of Mathematical Sciences and Applications (BIMSA), Beijing, China\\
\texttt{wangzhong@bimsa.cn}}

\begin{document}
\maketitle

\begin{abstract}
Multimedia decision support requires more than recognition; it requires
explicit state estimates that can be checked against rules, audited by humans,
and consumed by downstream decision logic. We present the FEncing Referee
Assistant (FERA), a pose-based framework for this setting, and study it
through foil fencing, where
decisions depend on fast bilateral motion and right-of-way rules. The
framework separates canonical participant tracking, kinematic tokenization,
calibrated temporal perception, a compact structured decision layer, and an
explanation-oriented retrieval interface. We also release an audited
benchmark with adjudicated labels and fixed folds for reproducible evaluation.
Under a shared protocol, a lightweight lifted-depth sidecar strengthens the
best graph-based perception model, while a compact structured classifier on
the fixed two-dimensional token stream reaches 0.624 accuracy and a 0.632
macro-averaged F1 score on the final Left / Right / None decision. The case
study supports a broader design lesson: keep the boundary between perception
and rule application explicit, preserve uncertainty, and choose the
perception front end according to the downstream operating point.
\end{abstract}


\section{Introduction}

Understanding human activities from video is a central problem in multimedia
analysis. In many applications, however, recognition is only the first step.
The harder systems problem is to convert uncertain visual evidence into
structured state estimates that can support symbolic rules, explanations, and
human-facing decisions. This appears in sports officiating, procedural
analysis, compliance monitoring, and other domains in which the desired output
is not merely ``what happened,'' but ``what decision follows from what
evidence.''

Foil fencing is a challenging and concrete instance of this broader problem.
Referees must track two fencers, interpret fast blade exchanges, and apply
right-of-way (priority) rules to decide who is awarded the point when hits
occur in close succession. Even at elite competitions, calls can be
controversial; in practice settings, trained officials may not be available.
While electronic scoring indicates \emph{when} a touch occurs, it does not
explain \emph{why} a point is awarded. The sport therefore offers a stringent
case study for multimedia decision support: perception is noisy, interaction is
bilateral, and the final output is explicitly rule grounded.

Recent work has begun to close this gap in other sports, for example through
multi-view video assistant referee systems and event spotting benchmarks for
soccer~\cite{held2023vars,held2024xvars,soccernet2024mvfoul}, as well as
pose-conditioned temporal models for skeleton-based action
recognition~\cite{bai2018tcn,xu2022vitpose}. These efforts highlight sports
officiating as an emerging AI application with a plausible path toward
deployment. At a systems level, however, the central question is not only how
well one recognizes actions, but how visual evidence should be represented so
that downstream decision modules can reason over it reliably. We study that
question in broadcast fencing video and use the domain as a concrete case
study for broader design choices about representation, uncertainty, and rule
grounding.

We treat fencing exchanges as a case-study setting in which video must be
translated into tokens suitable for downstream reasoning. Our immediate goal is
to predict, from monocular video, (i) the multi-label set of footwork and
blade actions active for each fencer over time and (ii) the instantaneous
blade-line position, one of the standard foil lines
$\{4,6,7,8,\text{shoulder}\}$. More broadly, we use this setting to study a
reusable systems pattern: represent each participant in a canonical view,
infer temporally localized action and state tokens with calibrated confidence,
and feed those tokens to a symbolic or language-based reasoner. To simplify
downstream processing and avoid fencer-specific logic, we extract per-fencer
sequences by applying the same single-fencer pipeline to the original clip and
to a horizontally flipped copy. This yields two time-aligned processed views
in which the target fencer always occupies the left side of the frame, while
detection, tracking, and feature extraction remain identical for both sides.

Our framework, the FEncing Referee Assistant (FERA), separates perception from
reasoning. First, 2D poses are converted into a 101-dimensional kinematic
feature sequence in a canonical single-fencer view; this base token can also
be augmented with a lightweight 24-dimensional z-only pose-lift sidecar.
Second, a temporal perception module (FERA-MBD) is trained on trimmed
exchanges and deployed on arbitrary-length pose tracks using dynamic temporal
windows to infer segment boundaries. We evaluate FERA-MBD with several
backbone families under a single fixed protocol: transformer variants,
graph-temporal models, compact pseudo-3D baselines, RGB fusion, and custom
hybrids. Finally, paired left/right predictions are converted into tokens and
consumed in two ways: by a compact structured decision layer that predicts the
final Left / Right / None outcome from a fixed 2D token stream, and by an
explanation-oriented interface (FERA-LM) that combines retrieved rule text
with structured action tokens to produce human-readable decision traces.

\paragraph{Contributions.}
\begin{itemize}[leftmargin=*,itemsep=2pt]
    \item We present FERA as a pose-based pipeline for studying
    rule-grounded multimedia decision support through a foil fencing case
    study, separating pose tracking, kinematic tokenization, calibrated
    temporal perception, and downstream rule application through an explicit
    token interface.
    \item We identify and evaluate concrete design choices for this setting:
    a canonical single-agent view shared across participants, dynamic temporal
    windowing with explicit no-consensus handling, and a token interface
    between perception and reasoning.
    \item We release a fencing benchmark and controlled empirical study built
    around this pipeline: 493 underlying clip IDs, 598 matched pose-view
    files, 1,800 cleaned labeled segments, a full accepted-subset second
    review with joint adjudication, and a fixed multilabel-stratified fold
    definition for reproducible backbone comparisons and ablations. A
    lightweight 24D z-only pose-lift sidecar strengthens the best graph model,
    broader graph-baseline comparisons do not displace lifted ST-GCN, a
    mixed-orientation ablation confirms the value of canonicalization, and a
    compact random-forest structured decision layer together with
    threshold-transfer and boundary-quality checks provides a quantitative
    downstream evaluation on top of the token stream.
\end{itemize}

\section{Related Work}

\subsection{AI for sports refereeing}

Several works have explored AI-assisted refereeing and decision support in
sports. In fencing, prior work has focused mainly on controlled footwork
datasets: FenceNet and related models report high accuracy on the Fencing
Footwork Dataset~\cite{zhu2022fencenet,footworkdataset}, which contains 652
videos from 10 fencers recorded under fixed camera angles and annotated with a
small set of footwork actions. While valuable for isolated motion analysis,
these datasets omit blade actions, opponent interactions, and real broadcast
conditions, limiting their applicability to foil referee decisions.

In soccer, Video Assistant Referee Systems (VARS) combine multi-view analysis
with temporal localization of fouls and offside
events~\cite{held2023vars,held2024xvars}, and benchmarks such as SoccerNet
target decision-making and explanation quality in realistic broadcast
footage~\cite{soccernet2024mvfoul}. Unlike SoccerNet-style event spotting,
which focuses on second-level events in multi-view soccer, our task operates
on single-view broadcast footage and requires frame-level multi-label
recognition and rule reasoning in a two-person weapon sport. FERA extends this
line of work to foil fencing, where exchanges are short, motion is fast and
mostly linear, contacts are point-based, and right-of-way depends on subtle
blade actions and timing.

\subsection{Pose-based action recognition}

Human pose estimation supports action recognition by mapping video to joint
trajectories. Bottom-up multi-person methods such as
OpenPose~\cite{cao2017openpose} introduced part affinity fields, while
top-down approaches combine person detection with single-person pose
estimators~\cite{xiao2018simple}. Transformer-based models like
ViTPose~\cite{xu2022vitpose} further improve accuracy via large-scale
pretraining. We adopt a standard detection--pose stack, using MMDetection and
MMPose~\cite{mmdetection2019,mmpose} with RTMDet and
RTMPose~\cite{rtmpose2023,rtmdet2023} for efficient fencer detection and 2D
keypoint estimation.

For temporal modeling, recurrent networks, temporal convolutional networks
(TCN)~\cite{bai2018tcn}, graph-convolutional models such as
ST-GCN, 2s-AGCN, and CTR-GCN~\cite{yan2018stgcn,shi2019twosagcn,chen2021ctrgcn},
and transformer-style pose models~\cite{zheng2021poseformer,zhao2023poseformerv2}
have all been explored for skeleton-based action recognition, with growing
interest in calibrated probabilities~\cite{guo2017calibration}. PoseC3D-style
pose heatmap baselines and RGB-pretrained video encoders such as
VideoMAE~\cite{duan2022posec3d,tong2022videomae} further show that visual
context can recover cues not present in raw joint coordinates. Existing
pose-based datasets, however, focus on everyday actions or sports without
handheld weapons and therefore ignore fine-grained bilateral interactions. In
fencing, foil blades move quickly and are often blurred or absent in
low-frame-rate broadcast footage, making explicit geometric blade
reconstruction unreliable.

This setting motivates models that infer blade-line implicitly from body
dynamics (e.g., wrist--elbow configuration, arm extension, timing) rather than
from direct blade detection. Our method follows this strategy by jointly
learning moves and blade-line from pose-derived kinematic features while also
keeping the evaluation protocol compatible with stronger modern skeleton
baselines.

Beyond sports-specific baselines, our design is also adjacent to broader
neuro-symbolic and programmatic multimedia pipelines in which perception is
forced to expose explicit intermediate states before any rule application.
FERA follows that philosophy deliberately: instead of asking one end-to-end
model to internalize both motion perception and foil rules, we keep the token
boundary explicit so that downstream decision modules can inspect move timing,
blade-line estimates, and confidence values directly.

FERA is also adjacent to temporal action segmentation in the narrower sense
that deployment quality depends not only on label recognition, but also on
where segment boundaries are placed and whether a system over-segments or
skips ambiguous intervals. Our inference procedure differs from dense
frame-label refinement pipelines because it emits multi-label windows and
blade states rather than per-frame action strings, but the same concern with
boundary fidelity motivates the boundary-quality analysis reported in
Section~\ref{sec:structured_e2e}.

This emphasis also connects FERA to uncertainty-aware decision support. The
goal is not only to maximize recognition scores, but to preserve a usable
confidence signal for later operating-point selection, abstention, and human
audit. That is why the paper reports calibration metrics alongside F1, and
why the structured decision layer is analyzed as a consumer of calibrated
tokens rather than as a separate black-box classifier.

\section{Foil Fencing Case Study Dataset}
\label{sec:data}

\subsection{Source videos and clip extraction}

The FERA dataset is constructed from professional foil bouts recorded at three
Grand Prix competitions (720p, 25\,fps), covering approximately 130
competitors. For each match, we locate candidate exchange segments by
scanning for changes in the electronic scoreboard and hit lights. Static
overlays for score and hit indicators are localized once per event, and color
and digit recognition are used to detect hits and score changes. These
signals are used only to align clips with official referee decisions.

Each match is segmented into short clips containing the final seconds
preceding a touch. We retain clips with clear views of the piste and both
fencers, and discard segments where camera cuts or occlusions prevent reliable
tracking.

All source videos are publicly available competition recordings; we release only
anonymized pose features, labels, and code.

\subsection{Annotation protocol}
\label{sec:annotation}

Each clip has a unique identifier and is converted to a canonical left-side
view: in the original clip the annotator labels the left fencer; in a
horizontally flipped copy, the original right fencer is mirrored onto the left
side and labeled there. All first-pass annotations were produced by the first
author, a competitive foil fencer with nine years of experience, using slowed
replay and frame-by-frame scrubbing. A second review pass then focused on pose
quality: every labeled interval was reopened with the extracted joints
overlaid, and intervals with tracking drift, wrong-target assignment, or
unusable joint sequences were rejected or marked as junk.

To strengthen label quality beyond the original pass, a second annotator
reviewed every accepted segment before the final pose-validity filter,
covering 1,103 accepted single-fencer segments. This produced 1,022 immediate
agreements and 81 disagreements. Those disagreements were adjudicated jointly
into three outcomes: keep the original label, revise the move label set, or
reject the segment entirely. In the final labels, 75 segments received
move-label changes and 6 segments were removed from the accepted pool.
Blade-position labels were unchanged during this adjudication pass. The
experiments in Sections~\ref{sec:experiments} and the released fold definitions
use these adjudicated labels. Disagreements were concentrated in
interaction-heavy moves such as beat, parry, and wait, while common locomotion
labels such as step forward, step backward, and lunge remained highly
consistent.

Annotations were created with a custom internal tool that supports
frame-by-frame review and pose-overlay inspection. The annotator assigns one or
more actions from the 11-class vocabulary in Table~\ref{tab:dist}(a) to each
continuous interval together with a single blade-line position in
$\{4, 6, 7, 8, \text{shoulder}\}$. In standard foil convention, position~4
denotes high inside line, 6 high outside, 7 low inside, and 8 low outside.
Overlapping actions (e.g., beat plus step forward) receive all relevant
labels. Ambiguous exchanges are revisited by reopening the surrounding replay,
rechecking adjacent actions on both sides, and either committing to a
consistent intervalization or marking the segment as \emph{junk}. Here,
\emph{junk} denotes unusable intervals, typically caused by camera cuts,
severe occlusion, missing target-fencer tracking, or replay timing that
prevents a defensible action label.

The tool also stores an explicit \texttt{accept}/\texttt{reject} status for
each interval. We therefore report both the full cleaned label inventory and
the stricter accepted-and-pose-valid subset used for quantitative evaluation.
During adjudication, a previously accepted segment can remain unchanged, be
relabeled, or be converted to \texttt{reject}. When both sides are annotated,
one clip yields two single-fencer sequences with the same identifier, and
FERA-MBD is trained only on these canonical single-fencer sequences.

\begin{table}[t]
  \centering
  \caption{Dataset audit for the public release and the accepted-and-pose-valid evaluation subset.}
  \label{tab:audit}
  \scriptsize
  \begin{tabular}{lrr}
    \toprule
    \textbf{Statistic} & \textbf{Release} & \textbf{Eval. subset} \\
    \midrule
    Unique underlying clip IDs & 493 & 493 \\
    Matched labeled pose-view files & 598 & 598 \\
    Total labeled segments & 2,507 & 2,507 \\
    Junk segments removed & 656 & 656 \\
    Rejected segments & 0 & 727 \\
    Invalid-move segments removed & 51 & 51 \\
    Cleaned labeled segments & 1,800 & 1,045 \\
    Move instances & 2,583 & 1,456 \\
    \midrule
    Valid highfps segments & 610 & 471 \\
    Valid lowfps segments & 918 & 442 \\
    Valid lowfpsflip segments & 272 & 132 \\
    \bottomrule
  \end{tabular}
\end{table}

\begin{table}[t]
  \centering
  \caption{Distributions of move and blade-position labels in the cleaned public release.}
  \label{tab:dist}
  \begin{minipage}{0.48\columnwidth}
    \centering
    \scriptsize
    \textbf{(a) Moves}
    \vspace{2pt}

    \begin{tabular}{lrr}
      \toprule
      Move & Count & \% \\
      \midrule
      Step forward & 734 & 28.4 \\
      Step backward & 380 & 14.7 \\
      Half step forward & 146 & 5.7 \\
      Half step backward & 103 & 4.0 \\
      Lunge & 250 & 9.7 \\
      Fl\`eche & 14 & 0.5 \\
      Wait & 74 & 2.9 \\
      Parry & 96 & 3.7 \\
      Beat & 180 & 7.0 \\
      Counterattack & 86 & 3.3 \\
      Hit & 520 & 20.1 \\
      \bottomrule
    \end{tabular}
  \end{minipage}
  \hfill
  \begin{minipage}{0.48\columnwidth}
    \centering
    \scriptsize
    \textbf{(b) Blade positions}
    \vspace{2pt}

    \begin{tabular}{lrr}
      \toprule
      Position & Count & \% \\
      \midrule
      6 & 1162 & 64.6 \\
      8 & 413 & 22.9 \\
      7 & 106 & 5.9 \\
      4 & 87 & 4.8 \\
      Shoulder & 32 & 1.8 \\
      \bottomrule
    \end{tabular}
  \end{minipage}
\end{table}

\FloatBarrier

\subsection{Data distribution}

Across the full corpus, the three source directories contain 493 unique
underlying clip IDs and 598 matched labeled pose-view files. After removing
656 junk segments and 51 invalid move segments, the cleaned release contains
1,800 labeled single-fencer segments with 2,583 move instances
(1.435 moves/segment on average). Source-wise, this corresponds to 610 valid
segments from \texttt{highfps}, 918 from \texttt{lowfps}, and 272 from
\texttt{lowfpsflip}. For the perception study we apply a stricter evaluation
subset: only adjudicated \texttt{accept}-status segments that also pass
pose-bound validation are retained, leaving 1,045 trainable segments.
Table~\ref{tab:audit} summarizes the audit counts, while
Table~\ref{tab:dist} reports the cleaned release label distributions.

Perception experiments use 5-fold multilabel-stratified cross-validation, with
splits defined at the clip level so that frames from the same clip never
appear in both training and test sets. We do not enforce disjoint athletes because the broadcast metadata
does not consistently identify the fencers. To partly address this limitation,
we also report grouped match-pool evaluation in
Section~\ref{sec:robustness} and for the structured decision baseline. In that
split, clips are grouped by the match identifier inferred from the filename,
and all clips from the same match remain in the same fold. The
\texttt{lowfps} and \texttt{lowfpsflip} directories share a common match pool.
We do not report a stricter prefix-only split as a main benchmark because the
available filename prefixes are too coarse and too uneven to serve as a
credible athlete- or competition-disjoint protocol.

\subsection{Training-time augmentation}

After splitting into folds, we apply data augmentation to improve robustness.
Augmentations include temporal jitter, additive noise, small rotations, and
scaling. Minority move classes are oversampled, and blade-line imbalance is
handled through loss weighting. At evaluation time, the main comparison
protocol tunes per-class decision thresholds independently on each validation
fold. The legacy \texttt{newversionv3} anchor retained in the repository also
applies per-class temperature scaling, but the backbone comparisons reported in
this paper use the fixed SMC evaluation path with raw validation logits and
per-fold threshold selection.

\section{Method}
\label{sec:method}

\begin{figure}[!tb]
  \centering
  \IfFileExists{pipeline.png}{\includegraphics[width=\columnwidth]{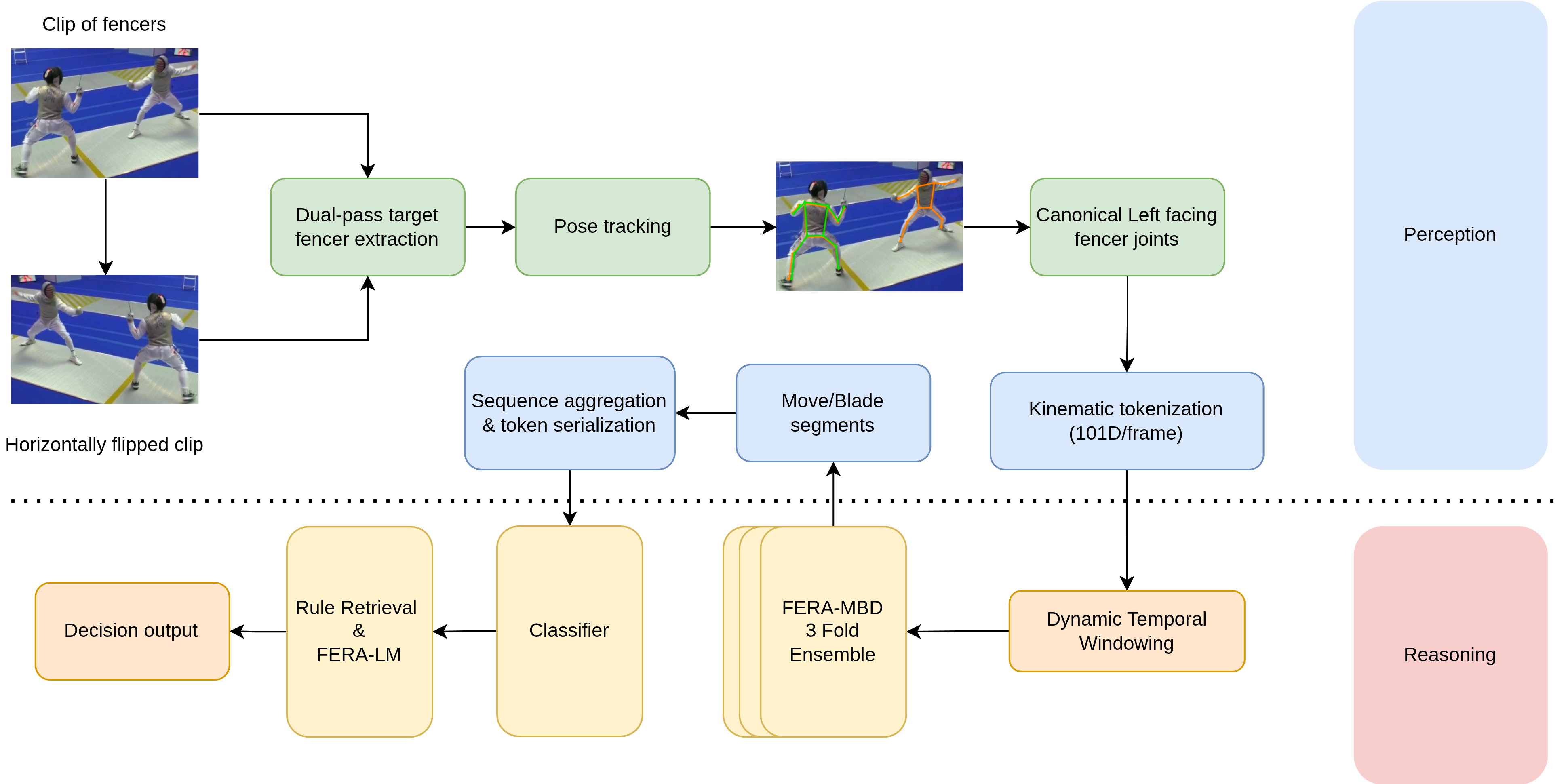}}{\fbox{\parbox[c][0.18\textheight][c]{0.95\columnwidth}{\centering Missing file: \texttt{pipeline.png}}}}
  \caption{FERA pipeline from broadcast video to participant-centric tokens, optional lifted-depth augmentation, a structured decision layer, and an explanation-oriented rule interface. The framework isolates the perception stack from downstream decision logic through an explicit token boundary.}
  \label{fig:pipeline}
\end{figure}

The FERA framework follows a modular perception-to-reasoning pipeline
(Fig.~\ref{fig:pipeline}). Starting from monocular broadcast video, the system
first detects and tracks individual fencers, extracts 2D human poses, and
converts them into a canonical single-fencer representation. These pose
sequences are mapped to kinematic feature tokens and processed by a temporal
model (FERA-MBD) to recognize multi-label actions and blade-line positions
over time. Finally, the structured predictions from both fencers are
aggregated into symbolic tokens and summarized structured features that feed
two downstream consumers: a lightweight decision layer for the final
Left / Right / None outcome, and a language-based reasoning module (FERA-LM)
that produces a referee-style explanation.

The framework separates perception from reasoning at an explicit token
boundary. Canonicalization, temporal perception, confidence estimation, and
sequence serialization are fixed system components. The foil-specific content
appears only in the move vocabulary, the blade-line labels, the hit semantics,
and the rule text retrieved by FERA-LM.

\subsection{Pose tracking}

We obtain per-fencer pose tracks using a top-down detect--pose--track pipeline
(detect persons, estimate single-person pose, then associate tracks over
time). RTMDet-Tiny (MMDetection) detects person bounding boxes and RTMPose-M
(MMPose) extracts 2D joints for each detected person~\cite{mmdetection2019,rtmdet2023,mmpose,rtmpose2023}.
Detections are linked over time using Norfair~\cite{norfair_repo,norfair_docs}
with a distance combining box IoU and centroid distance. If an ID is lost, it
is recovered by matching new detections to the most recent boxes. Joint
trajectories are smoothed with an exponential moving average to reduce
short-term jitter while preserving fast fencing motion.

We then select the target fencer track in each view with lightweight filtering
and scoring. In the first few frames, we suppress obvious non-target detections
by retaining only boxes whose vertical centers fall within a central action
band and whose areas exceed a minimum fraction of the frame, removing small
background detections and spectators near the image borders. During tracking,
we apply a looser area filter and rely on temporal continuity.

To further reject non-fencers (e.g., referees), we apply a fencer-candidate test
based on detector confidence, box size, and pose geometry. Very small or
low-score tracks are dropped; when poses are available, each track is
associated with its most-overlapping pose and candidates with joints
concentrated near the bottom of the frame are discarded, which commonly
corresponds to referees off the piste. Among remaining tracks, a foreground
score favors large, upright people on the left side and penalizes boxes near
the top/bottom borders; scores are accumulated over the first 15 frames and the
best track is selected.

This procedure is run on the original clip and on a horizontally flipped copy,
yielding two single-fencer tracks that share the same frame indices and are
naturally time-aligned. In both processed views, the target fencer appears on
the left side of the frame. Finally, tracks are mapped to a shared canonical
coordinate system so that their geometry is consistent with the annotation
convention in Section~\ref{sec:annotation}.

\subsection{Kinematic representation}

After removing head keypoints, each frame is represented by 12 body joints
(shoulders, elbows, wrists, hips, knees, ankles) in the shared canonical
coordinate system. Upstream processing normalizes these keypoints for
translation and approximate scale, and missing or unreliable joints are set to
zero.

From these normalized joints and precomputed joint angles, we build a
49-dimensional static descriptor per frame:
\begin{itemize}[leftmargin=*,itemsep=1pt]
    \item \textbf{Normalized joints (24D):} flattened $(x,y)$ coordinates for the 12 joints.
    \item \textbf{Center of mass (2D):} mean of all nonzero joint coordinates.
    \item \textbf{Pairwise distances (11D):} distances for a fixed set of segments and widths (upper/lower arm and leg segments, shoulder and hip width, and hand separation).
    \item \textbf{Joint angles (4D):} left/right elbow and left/right knee angles.
    \item \textbf{Torso orientation (2D):} sine and cosine of the angle of the shoulder--hip axis.
    \item \textbf{Arm extension (6D):} for each arm, the shoulder--wrist distance and the corresponding unit direction vector (2D).
\end{itemize}

To capture motion, we append first- and second-order finite differences for
(i) the 24 normalized joint coordinates and (ii) the 2D center of mass,
contributing 52 additional dimensions. The base per-frame descriptor
therefore has $49$ static + $52$ motion = $101$ features. In the
3D-augmented branch, we complement this base token with a compact 24D depth
sidecar obtained from an off-the-shelf VideoPose3D lifter in MMPose. Starting
from the raw RTMPose detections associated with the target fencer, we lift a
27-frame pose sequence with the pretrained Human3.6M model, retain the same
12 body joints used by the 2D representation, normalize the recovered body
around the hip midpoint and a shoulder/hip-width scale, and keep only the
per-joint depth coordinates and their framewise depth velocities. The result
is a lightweight relative-depth cue rather than a metric 3D reconstruction
tailored to fencing broadcast video. The final descriptor therefore has 125
dimensions for dense-token models and exposes the lifted component as an
explicit 24D auxiliary block for graph-temporal models that already consume
the 2D joints directly.

\subsection{Temporal windowing}

FERA-MBD is trained on trimmed single-fencer segments in which all active moves
share the same temporal boundary: if a beat occurs during a step forward, both
labels are attached to the same frame interval. At inference time, the model
therefore predicts, for each input sequence, a multi-label set of moves that
are assumed to share a single start and end frame, together with one blade-line
class.

In deployment, FERA-MBD must operate on per-fencer pose tracks of arbitrary
length, where segment boundaries are unknown. We use a dynamic temporal
windowing strategy to discover these boundaries. Given a tracked pose sequence
for one fencer (25\,fps), we start from frame $s$ with a short window and
gradually expand it up to a maximum of 60 frames (about 2.4\,s). For each
candidate window we evaluate a 3-model FERA-MBD ensemble, where each model is
trained on a different training fold and produces 11 move probabilities and a
5-way blade distribution. Missing joints are zero-filled upstream, but frame
order is preserved: we never delete frames inside the dynamic window. This
windowing-and-ensemble procedure is used only for deployment on untrimmed
tracks; cross-validation metrics in Section~\ref{sec:experiments} use one
model per fold on trimmed segments.

For a move class $c$ to be accepted in the current window, at least two out of
three models must predict $p_c > \tau_c$, where $\tau_c$ is the per-class
threshold tuned on validation data. The blade-line class for the window is
chosen by majority vote over the three models' blade predictions. If at least
one move class is accepted, we record the current window $[s,e]$ as a segment
with a multi-label move set and a single blade-line label and resume from
frame $e{+}1$. If no move is accepted by the maximum window length, we emit no
segment for that start index, advance to frame $s{+}1$, and repeat. This
yields an ordered sequence of non-overlapping segments; co-occurring moves
(e.g., step forward + beat) share the same temporal window and thus the same
boundaries.

\begin{figure}[!tb]
  \centering
  \fbox{\parbox{0.96\columnwidth}{
  \textbf{Algorithm 1: Dynamic temporal window inference}\par
  \vspace{2pt}
  1. Initialize the current start frame $s$.\par
  2. Expand the candidate window $[s,e]$ from a minimum length up to 60 frames.\par
  3. For each window, run the 3-model ensemble and accept move $c$ only if at least two models satisfy $p_c>\tau_c$.\par
  4. If one or more moves are accepted, emit segment $[s,e]$, assign the blade line by majority vote, and continue from $e{+}1$.\par
  5. If no move is accepted by the maximum length, emit nothing and continue from $s{+}1$.
  }}
  \caption{Exact no-consensus handling used for deployment on untrimmed pose tracks.}
  \label{fig:dynamic_algorithm}
\end{figure}

Processing each bout twice, once in its original orientation and once after
horizontal flipping, yields two canonical single-fencer views with the same
global frame indices. Downstream comparison therefore requires no explicit
multi-person pose association, identity tracking, or cross-person
synchronization.

\subsection{FERA-MBD}

FERA-MBD denotes the perception module that maps a trimmed single-fencer
segment to 11 move logits and 5 blade-line logits. The base input is the
101-dimensional kinematic sequence described above; in the depth-augmented
branch we append the 24D z-only sidecar. What changes across experiments is
therefore both the temporal backbone and, in the depth-augmented branch, whether that
backbone consumes the lifted-depth augmentation.

Formally, given a segment of length $T$ with per-frame descriptors
$x_t \in \mathbb{R}^{D}$, where $D{=}101$ for the 2D-only branch and
$D{=}125$ for the z-only lifted branch, a backbone $f_\theta$ produces a
segment-level
representation
\begin{equation}
h = f_\theta(x_1,\dots,x_T).
\end{equation}
Two prediction heads then output multi-label move logits and blade-line logits:
\begin{equation}
\ell^{(\text{move})} = W_m h + b_m, \qquad
\ell^{(\text{blade})} = W_b h + b_b.
\end{equation}

Let $p_c=\sigma(\ell_c^{(\text{move})})$. Move logits use weighted binary
cross-entropy
\begin{equation}
\mathcal{L}_{\text{move}}
= -\sum_{c=1}^{11} w_c \left[y_c \log p_c + (1-y_c)\log(1-p_c)\right],
\label{eq:moveloss}
\end{equation}
while blade-line prediction uses class-weighted cross-entropy
\begin{equation}
\mathcal{L}_{\text{blade}}
= -\sum_{k=1}^{5} \alpha_k\,\mathbf{1}[k=y^{\text{blade}}]
  \log\frac{e^{\ell_k^{(\text{blade})}}}{\sum_j e^{\ell_j^{(\text{blade})}}}.
\label{eq:bladeloss}
\end{equation}
The total loss is
\begin{equation}
\mathcal{L}
= \mathcal{L}_{\text{move}}
+ \lambda_{\text{blade}} \mathcal{L}_{\text{blade}},
\label{eq:totalloss}
\end{equation}
with $\lambda_{\text{blade}}$ tuned in the ablation study.
At evaluation time we also tune per-class decision thresholds on the
validation fold and report calibration metrics (ECE, MCE, and Brier) in
addition to F1.

We evaluate several backbone families under the same folds, labels, and
thresholds:
\begin{itemize}[leftmargin=*,itemsep=1pt]
    \item \textbf{Transformer variants.} We test FERA-Attn, FERA-Mask, and
    FERA-CLS. These share the same canonical kinematic input stream but differ
    in pooling strategy and treatment of missingness. FERA-Attn uses
    attention pooling, FERA-Mask adds explicit missingness cues, and FERA-CLS
    uses a learned classification token.
    \item \textbf{Graph-temporal models}: ST-GCN and CTR-GCN-style networks
    operating on the recovered 12-joint pose stream while retaining the same
    move/blade supervision.
    \item \textbf{Additional comparators.} We also test PoseFormerV2, a
    compact PoseC3D-style baseline, pose+RGB fusion, and two custom hybrids.
\end{itemize}

All models share the same segment-level supervision and fixed
multilabel-stratified folds, so differences in Section~\ref{sec:experiments}
reflect backbone choice and a small number of controlled training ablations
rather than changes in data or evaluation protocol.

\subsection{Sequence aggregation}

FERA-MBD produces two ordered lists of single-fencer segments,
\[
S_i^{(L)} = (s_i, e_i, M_i^{(L)}, B_i^{(L)}), \quad
S_j^{(R)} = (s_j, e_j, M_j^{(R)}, B_j^{(R)}),
\]
where $[s,e]$ denotes the frame range, $M$ the multi-label move set, and $B$
the blade-line class. Because the flipped clip is a horizontal mirror of the
original, both lists share the same global frame indices and are naturally
time-aligned.

For each fencer, segments are serialized into a compact representation
containing start/end frames, move identifiers and names, and blade-line names,
and formatted into a textual script in temporal order. From scoreboard signals
(Section~\ref{sec:data}), we derive one hit label for each side: no hit,
off-target, or on-target. These hit labels are prepended to the script.

The paired left/right scripts and hit labels are concatenated into a single
\emph{move sequence} string. We embed this string into a fixed-length dense
vector using a sentence-transformer and perform semantic retrieval over a small
rule database using FAISS~\cite{faiss2021}. The top-$k$ retrieved snippets are
then appended to the prompt used by the explanation stage. The same paired
segment lists are also summarized into a structured feature vector containing
segment counts, move confidences, earliest-action cues, blade-line summaries,
pairwise left/right contrasts, hit values, and an optional handcrafted
exchange-interaction summary. This serialized-plus-structured representation is
the contract between perception and reasoning: FERA-MBD outputs only move,
blade-line, and timing tokens, and downstream consumers operate on those
tokens rather than on raw video.

\subsection{Structured decision layer}

Alongside the language-based interface, we evaluate a compact decision module
that operates directly on the aggregated token stream. For each exchange, the
left and right segment lists are summarized into a fixed feature vector with
per-side segment counts, per-move confidence and first-occurrence statistics,
blade-line summaries, hit metadata, and an optional handcrafted
exchange-interaction summary. We then
train lightweight multiclass classifiers under grouped match-pool
cross-validation to predict the final priority outcome \{Left, Right, None\}.
This module serves as the paper's quantitative end-to-end baseline, while
FERA-LM remains a qualitative explanation consumer of the same token stream.

\subsection{FERA-LM}

FERA-LM is an explanation interface built on top of the
serialized left/right token stream. A fixed prompt defines the token format,
appends retrieved rule snippets, and requests a single priority decision
(Left / Right / None) together with a concise textual justification. In the
current system we use an off-the-shelf instruction-tuned language model
without fine-tuning and treat the module as a demonstration of explanation
feasibility rather than as the paper's primary quantitative contribution. In
this pipeline, final quantitative decisions are reported through the
structured decision layer above, while FERA-LM is used to verbalize the same
token stream and retrieved rule snippets for human inspection.

\section{Case Study Evaluation}
\label{sec:experiments}

\subsection{Evaluation protocol}

We evaluate move and blade recognition using 5-fold multilabel-stratified
cross-validation on trimmed single-fencer segments. For each fold we construct
training and held-out sets with similar label distributions and apply the data
processing and augmentation pipeline described in Sections~\ref{sec:data} and
\ref{sec:method}. Dynamic temporal windows and the 3-model ensemble are used
only when deploying on continuous pose tracks, not for computing
cross-validation metrics. To avoid silent protocol drift across model
families and ablations, we materialize and release the exact fold assignment
used in all runs. The public corpus audit covers
1,800 cleaned segments, while the perception experiments in this section use
the stricter subset of 1,045 adjudicated \texttt{accept}-status segments that
also pass pose-bound validation.

Unless a row is explicitly marked as 2D in Table~\ref{tab:ablation}, the main
perception results below refer to the 3D-augmented branch that appends
the 24D z-only lift sidecar to the base 101D token stream. The downstream
structured decision layer in Section~\ref{sec:structured_e2e} is still reported
on the fixed 2D ST-GCN token stream. A direct 2D-vs-3D token-stream
comparison in Section~\ref{sec:structured_e2e} explains why: the lifted 3D
branch slightly improves trimmed recognition and boundary alignment, but the
resulting exchange summaries reduce final-decision macro-F1 and None handling
for the current lightweight structured classifier.

To make the protocol fully reproducible, each fold uses its own validation
split to tune class thresholds after training; the released SMC runs in this
paper do not apply extra temperature scaling beyond that validation-time
threshold selection. For deployment on untrimmed tracks, the default dynamic
window uses a minimum length of 8 frames, a maximum length of 60 frames, a
2-of-3 consensus rule across fold-specific checkpoints, and zero threshold
shift relative to the stored validation thresholds. In the grouped
match-pool protocol, all clips from the same inferred match identifier stay in
the same fold, with \texttt{lowfps} and \texttt{lowfpsflip} sharing the same
pool. Thresholds are always chosen only on the validation split inside each
training fold; the held-out fold is untouched until final evaluation, and the
threshold-transfer analysis in Section~\ref{sec:structured_e2e} compares these
fold-specific thresholds against shared and fixed alternatives.

\subsection{Backbone comparison}

This section identifies which perception backbones best support the fencing
task under one fixed protocol. For each run we report move macro-F1, blade
macro-F1, combined macro-F1 across all 16 labels, micro-F1, weighted-F1,
Hamming loss, and calibration metrics (ECE, MCE, and Brier
score)~\cite{guo2017calibration,brier1950}. Unless noted otherwise, all
models use the same folds, class definitions, threshold tuning, and
evaluation code.

Table~\ref{tab:model_comparison} summarizes the shared 3D-augmented backbone
comparison, using one selected operating point per family together with four
additional compact graph baselines that broaden the comparison.

Four patterns stand out. First, graph-temporal models are the strongest family
for move detection in this branch: lifted ST-GCN with no temporal
jitter reaches 0.517 move macro-F1, 0.450 blade macro-F1, and 0.496 combined
macro-F1 while remaining small and well calibrated. The additional
2s-AGCN-style, MS-G3D-style, ResGCN-style, and PartAtt-style baselines do not
displace it; all four remain between 0.394 and 0.408 combined macro-F1 under
the same folds, thresholds, and lifted 24D sidecar. Second, PoseC3D-style
benefits from the same depth augmentation and becomes the strongest
blade-line model among the selected operating points (0.470 blade macro-F1)
while also achieving the lowest ECE (0.094). Third, the in-house transformer
family remains competitive but no longer leads on calibration overall: the
strongest transformer operating point is FERA-CLS with no temporal jitter
(0.465 move macro-F1, 0.461 combined macro-F1), while the no-sampler variant
remains the calibration-focused transformer operating point in the ablation
study. Fourth, the retrained pose+RGB fusion model and the custom hybrids
still remain below the strongest lifted-pose baselines, suggesting that added
architectural complexity does not help in this data regime.

\begin{table*}[t]
  \centering
  \caption{Backbone comparison with the shared 24D z-only pose-lift sidecar. Additional compact graph baselines are trained under the same folds, labels, and thresholds. Values are mean $\pm$ std over 5 folds.}
  \label{tab:model_comparison}
  \small
  \setlength{\tabcolsep}{4pt}
  \resizebox{\textwidth}{!}{%
  \begin{tabular}{lrrrrrr}
    \toprule
    \textbf{Model} & \textbf{Params} & \textbf{Move Macro-F1} $\uparrow$ & \textbf{Blade Macro-F1} $\uparrow$ & \textbf{Combined Macro-F1} $\uparrow$ & \textbf{ECE} $\downarrow$ & \textbf{Brier} $\downarrow$ \\
    \midrule
    ST-GCN & 115,984 & \textbf{0.517$\pm$0.008} & 0.450$\pm$0.034 & \textbf{0.496$\pm$0.010} & 0.097$\pm$0.004 & \textbf{0.092$\pm$0.003} \\
    CTR-GCN & 116,416 & 0.415$\pm$0.028 & 0.413$\pm$0.020 & 0.414$\pm$0.024 & 0.131$\pm$0.008 & 0.117$\pm$0.003 \\
    2s-AGCN-style & 525,040 & 0.420$\pm$0.017 & 0.336$\pm$0.030 & 0.394$\pm$0.016 & 0.125$\pm$0.021 & 0.120$\pm$0.009 \\
    MS-G3D-style & 366,112 & 0.419$\pm$0.018 & 0.379$\pm$0.035 & 0.406$\pm$0.016 & 0.119$\pm$0.010 & 0.111$\pm$0.003 \\
    ResGCN-style & 190,768 & 0.417$\pm$0.010 & 0.357$\pm$0.033 & 0.398$\pm$0.013 & 0.128$\pm$0.022 & 0.117$\pm$0.007 \\
    PartAtt-style & 255,329 & 0.423$\pm$0.013 & 0.376$\pm$0.032 & 0.408$\pm$0.018 & 0.122$\pm$0.020 & 0.115$\pm$0.008 \\
    PoseFormerV2 & 603,408 & 0.413$\pm$0.027 & 0.394$\pm$0.047 & 0.407$\pm$0.029 & 0.114$\pm$0.011 & 0.114$\pm$0.006 \\
    PoseC3D-style & \textbf{78,832} & 0.466$\pm$0.024 & \textbf{0.470$\pm$0.038} & 0.467$\pm$0.020 & \textbf{0.094$\pm$0.006} & 0.096$\pm$0.004 \\
    FERA-Attn & 613,137 & 0.442$\pm$0.016 & 0.451$\pm$0.023 & 0.445$\pm$0.015 & 0.111$\pm$0.010 & 0.110$\pm$0.006 \\
    FERA-Mask & 614,673 & 0.448$\pm$0.014 & 0.421$\pm$0.044 & 0.439$\pm$0.021 & 0.112$\pm$0.007 & 0.110$\pm$0.003 \\
    FERA-CLS & 613,265 & 0.465$\pm$0.016 & 0.453$\pm$0.026 & 0.461$\pm$0.015 & 0.103$\pm$0.007 & 0.105$\pm$0.005 \\
    Pose+RGB Fusion & 768,593 & 0.418$\pm$0.015 & 0.412$\pm$0.057 & 0.416$\pm$0.019 & 0.149$\pm$0.022 & 0.133$\pm$0.009 \\
    Hybrid v2 & 786,672 & 0.442$\pm$0.013 & 0.416$\pm$0.045 & 0.434$\pm$0.009 & 0.167$\pm$0.037 & 0.134$\pm$0.017 \\
    \bottomrule
  \end{tabular}}
\end{table*}

\begin{figure}[!tb]
  \centering
  \IfFileExists{model_tradeoff_scatter.pdf}{\includegraphics[width=\columnwidth]{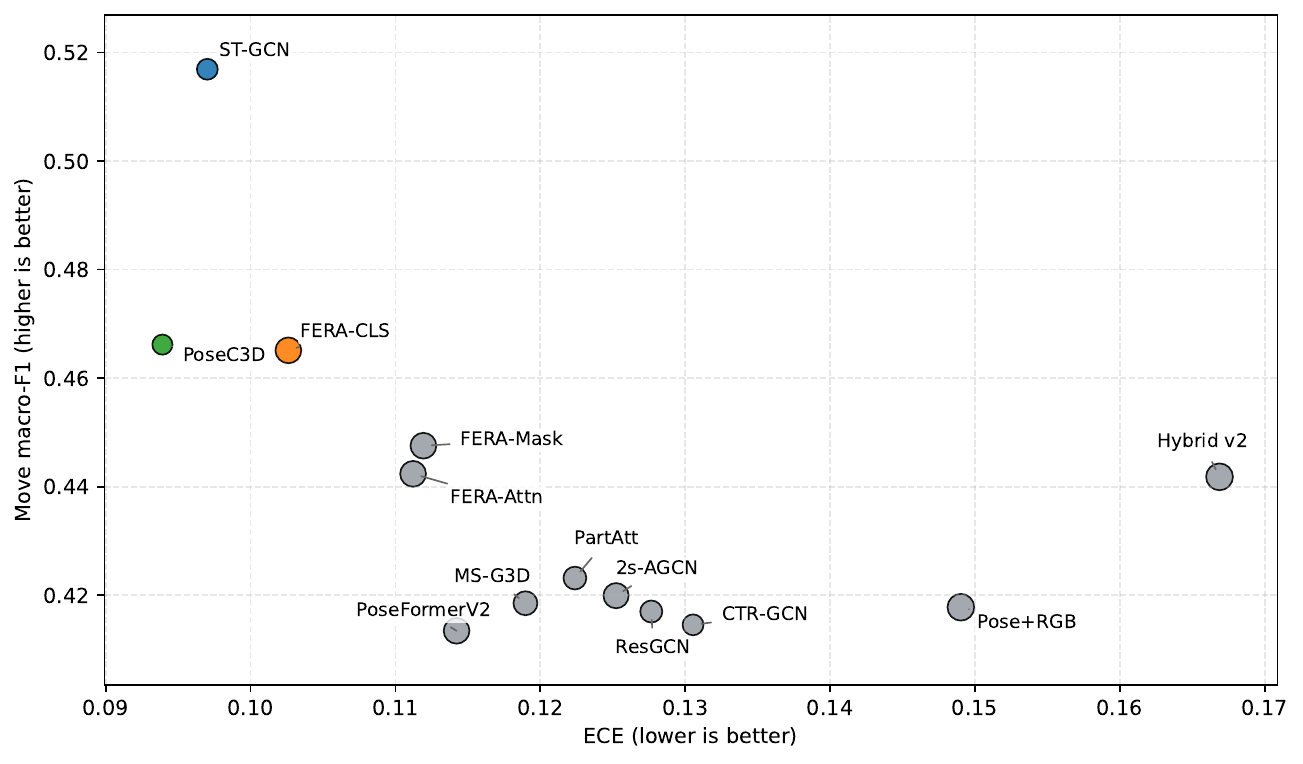}}{\fbox{\parbox[c][0.18\textheight][c]{0.95\columnwidth}{\centering Missing file: \texttt{model_tradeoff_scatter.pdf}}}}
  \caption{Move macro-F1 versus ECE for the 24D z-only 3D-augmented operating points. The lifted ST-GCN model now occupies the strongest move-sensitive region, while PoseC3D-style and the transformer variants offer alternative calibration trade-offs.}
  \label{fig:tradeoff_scatter}
\end{figure}

\begin{figure}[!tb]
  \centering
  \IfFileExists{calibration_summary.pdf}{\includegraphics[width=\columnwidth]{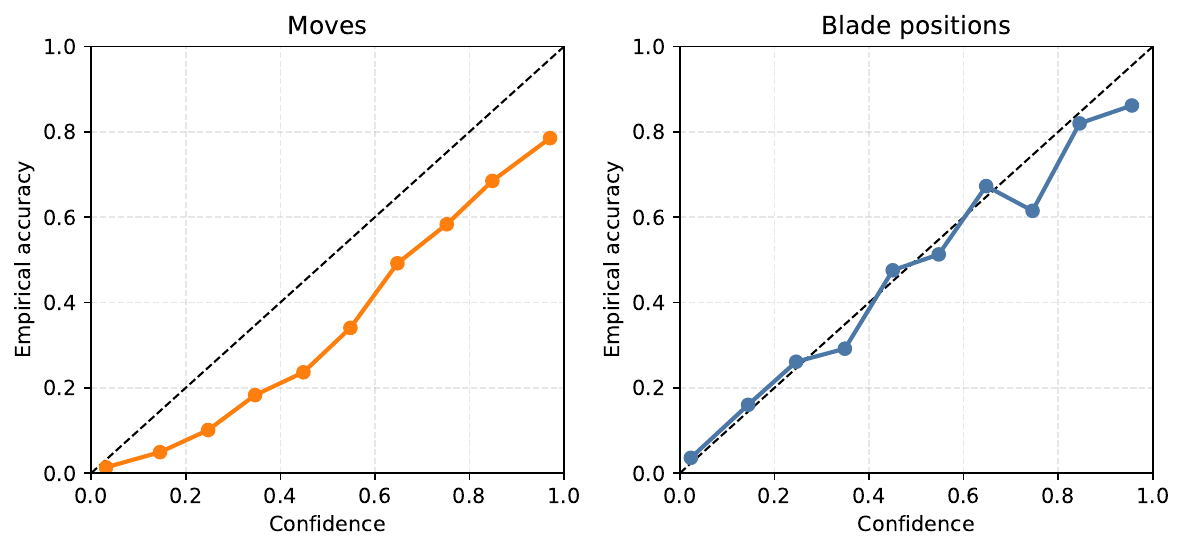}}{\fbox{\parbox[c][0.24\textheight][c]{0.95\columnwidth}{\centering Missing file: \texttt{calibration_summary.pdf}}}}
  \caption{Calibration diagnostics for the lifted ST-GCN no-jitter operating point. Reliability curves are shown separately for move labels and blade-line labels to complement the aggregate ECE and Brier scores in Table~\ref{tab:model_comparison}.}
  \label{fig:calibration}
\end{figure}

Figure~\ref{fig:tradeoff_scatter} visualizes all tested operating points from
Table~\ref{tab:model_comparison}, making the move-versus-calibration trade-off
visible at a glance across both the main backbones and the additional graph
baselines.
Figure~\ref{fig:calibration} complements the table with the aggregate move and
blade reliability curves for the lifted ST-GCN operating point.

\subsection{Design ablations}

We then reran a focused set of ablations to answer four practical questions:
how much the 24D lift helps over the original 2D token, whether
canonicalization still matters once depth cues are available, how
training-policy changes interact with the new sidecar, and whether the lift
helps other backbone families in the same way. Table~\ref{tab:ablation}
summarizes the most informative variants. These runs yield four especially
useful findings.

\noindent\textbf{The 24D z-only lift sidecar is now the decisive ST-GCN
improvement.}
Relative to the original 2D no-jitter ST-GCN baseline, the lifted no-jitter
model improves move macro-F1 from 0.491 to 0.517, blade macro-F1 from 0.438
to 0.450, and combined macro-F1 from 0.475 to 0.496 while also improving ECE
from 0.112 to 0.097 and Brier from 0.098 to 0.092. This is the main reason we
promote the lifted branch to the paper's primary perception benchmark.

\noindent\textbf{Canonicalization contributed measurable signal, not just
engineering convenience.}
The mixed-orientation ablation keeps the same lifted ST-GCN no-jitter
backbone, labels, folds, and thresholds, but trains on single-fencer
sequences in their source orientation instead of mapping both sides into the
shared left-oriented layout. Move macro-F1 falls from 0.517 to 0.489,
combined macro-F1 from 0.496 to 0.469, and both ECE and Brier worsen. This
shows that canonical dual-pass processing still reduces nuisance left/right
variation even after lightweight depth augmentation.

\noindent\textbf{Depth augmentation helps some families more than others.}
The same 24D sidecar helps FERA-Mask and PoseC3D-style substantially:
FERA-Mask long rises from 0.426 to 0.439 combined macro-F1, while
PoseC3D-style rises from 0.454 to 0.467 combined macro-F1 and also gains a
clear calibration benefit. FERA-CLS is more mixed. Its no-sampler operating
point slightly declines in combined macro-F1 (0.441 to 0.439) even though its
calibration improves, while the no-jitter 3D variant rises to 0.461 combined
macro-F1 and becomes the strongest in-house transformer.

\noindent\textbf{Sampling policy changed calibration more than architecture
alone.}
Within the transformer family, removing weighted sampling still improves
confidence quality even when it does not produce the highest move macro-F1.
This matters because FERA-MBD feeds downstream token streams with confidence
values rather than isolated class predictions. In the lifted branch,
FERA-CLS without weighted sampling remains the calibration-focused transformer
operating point (ECE 0.094, Brier 0.097), whereas the no-jitter variant is
the stronger recognition model.

\begin{table}[t]
  \centering
  \caption{Focused design ablations and grouped robustness checks. Values are mean $\pm$ std over 5 folds.}
  \label{tab:ablation}
  \scriptsize
  \textbf{(a) 2D-to-3D transition and focused ablations}\par
  \vspace{2pt}
  \setlength{\tabcolsep}{3pt}
  \resizebox{\columnwidth}{!}{%
  \begin{tabular}{lccccc}
    \toprule
    \textbf{Variant} & \textbf{Move} & \textbf{Blade} & \textbf{Combined} & \textbf{ECE} & \textbf{Brier} \\
    \midrule
    ST-GCN (2D no jitter) & 0.491$\pm$0.014 & 0.438$\pm$0.053 & 0.475$\pm$0.020 & 0.112$\pm$0.011 & 0.098$\pm$0.005 \\
    ST-GCN (lifted base) & 0.486$\pm$0.016 & 0.430$\pm$0.042 & 0.468$\pm$0.023 & 0.121$\pm$0.007 & 0.103$\pm$0.004 \\
    ST-GCN (lifted no jitter) & \textbf{0.517$\pm$0.008} & 0.450$\pm$0.034 & \textbf{0.496$\pm$0.010} & 0.097$\pm$0.004 & \textbf{0.092$\pm$0.003} \\
    ST-GCN (lifted no sampler) & 0.478$\pm$0.023 & 0.427$\pm$0.053 & 0.462$\pm$0.030 & 0.109$\pm$0.008 & 0.096$\pm$0.004 \\
    ST-GCN (lifted long budget) & 0.498$\pm$0.018 & 0.440$\pm$0.048 & 0.480$\pm$0.027 & 0.117$\pm$0.013 & 0.101$\pm$0.006 \\
    ST-GCN (lifted mixed orientation) & 0.489$\pm$0.015 & 0.424$\pm$0.044 & 0.469$\pm$0.022 & 0.113$\pm$0.008 & 0.099$\pm$0.006 \\
    \midrule
    FERA-CLS (2D no sampler) & 0.450$\pm$0.017 & 0.420$\pm$0.037 & 0.441$\pm$0.022 & 0.106$\pm$0.014 & 0.101$\pm$0.005 \\
    FERA-CLS (lifted no sampler) & 0.457$\pm$0.022 & 0.400$\pm$0.031 & 0.439$\pm$0.023 & 0.094$\pm$0.007 & 0.097$\pm$0.006 \\
    FERA-CLS (lifted no jitter) & 0.465$\pm$0.016 & 0.453$\pm$0.026 & 0.461$\pm$0.015 & 0.103$\pm$0.007 & 0.105$\pm$0.005 \\
    \midrule
    FERA-Mask (2D long) & 0.437$\pm$0.020 & 0.403$\pm$0.053 & 0.426$\pm$0.011 & 0.120$\pm$0.021 & 0.114$\pm$0.007 \\
    FERA-Mask (lifted long) & 0.448$\pm$0.014 & 0.421$\pm$0.044 & 0.439$\pm$0.021 & 0.112$\pm$0.007 & 0.110$\pm$0.003 \\
    \midrule
    PoseC3D-style (2D) & 0.445$\pm$0.028 & \textbf{0.474$\pm$0.056} & 0.454$\pm$0.030 & 0.109$\pm$0.012 & 0.101$\pm$0.007 \\
    PoseC3D-style (lifted) & 0.466$\pm$0.024 & 0.470$\pm$0.038 & 0.467$\pm$0.020 & \textbf{0.094$\pm$0.006} & 0.096$\pm$0.004 \\
    \bottomrule
  \end{tabular}}
  \par\medskip
  \textbf{(b) Match-grouped robustness}\par
  \vspace{2pt}
  \resizebox{\columnwidth}{!}{%
  \begin{tabular}{lcccccc}
    \toprule
    \textbf{Model} & \multicolumn{2}{c}{\textbf{Move}} & \multicolumn{2}{c}{\textbf{Blade}} & \multicolumn{2}{c}{\textbf{Combined}} \\
    \cmidrule(lr){2-3} \cmidrule(lr){4-5} \cmidrule(lr){6-7}
    & Canonical & Grouped & Canonical & Grouped & Canonical & Grouped \\
    \midrule
    ST-GCN & 0.517$\pm$0.008 & 0.513$\pm$0.029 & 0.450$\pm$0.034 & 0.432$\pm$0.046 & 0.496$\pm$0.010 & 0.488$\pm$0.020 \\
    FERA-CLS & 0.457$\pm$0.022 & 0.451$\pm$0.018 & 0.400$\pm$0.031 & 0.402$\pm$0.041 & 0.439$\pm$0.023 & 0.436$\pm$0.019 \\
    \bottomrule
  \end{tabular}}
\end{table}

\subsection{Grouped robustness test}
\label{sec:robustness}

To address the concern that clip-level folds may still mix exchanges from the
same bout, we constructed an additional grouped split for the two strongest
operating points: ST-GCN with no temporal jitter and FERA-CLS without weighted
sampling. The grouping key is the match identifier embedded in each filename,
so all clips from the same match stay in the same fold. Because
\texttt{lowfpsflip} is derived from the same underlying bouts as
\texttt{lowfps}, those two directories share the same match pool in this
grouped split.

The lower block of Table~\ref{tab:ablation} shows that the main
move-recognition conclusion is
stable under this stricter protocol. The lifted ST-GCN model changes from
0.517 to 0.513 move macro-F1 and from 0.496 to 0.488 combined macro-F1, while
the calibration-focused lifted FERA-CLS model changes from 0.457 to 0.451
move macro-F1 and from 0.439 to 0.436 combined macro-F1. This suggests that
the headline move-recognition performance is not being driven primarily by
same-match leakage. Blade macro-F1 drops modestly for lifted ST-GCN (0.450 to
0.432) and remains nearly unchanged for the lifted FERA-CLS no-sampler model
(0.400 to 0.402), indicating that the z-only sidecar does not destabilize the
match-grouped generalization story.

\subsection{Classwise behavior}

\begin{figure}[!tb]
  \centering
  \IfFileExists{per_class_f1.pdf}{\includegraphics[width=\columnwidth]{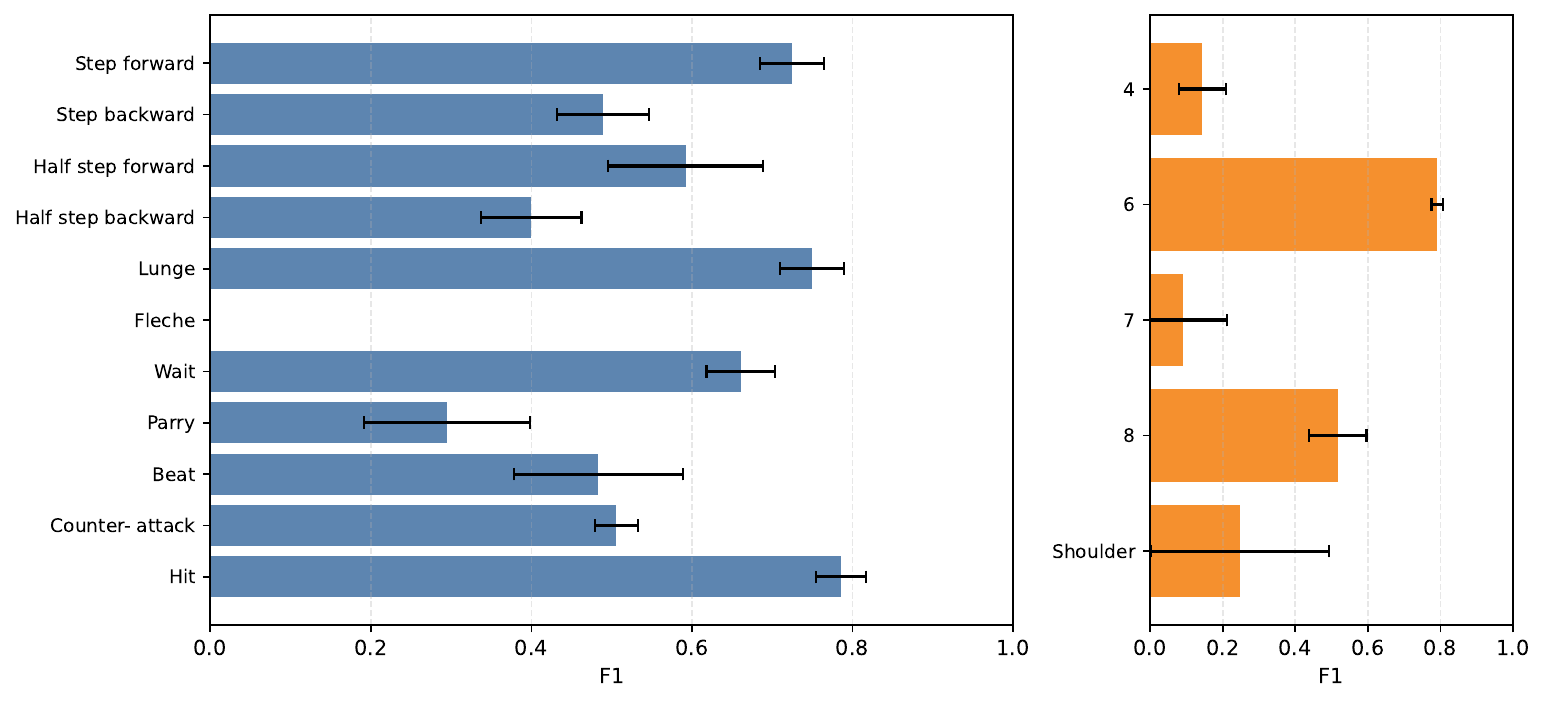}}{\fbox{\parbox[c][0.18\textheight][c]{0.95\columnwidth}{\centering Missing file: \texttt{per_class_f1.pdf}}}}
  \caption{Per-class F1 for the lifted ST-GCN no-jitter operating point, aggregated across folds. The remaining error is concentrated in rare and interaction-heavy classes rather than in common footwork categories.}
  \label{fig:per_class_f1}
\end{figure}

Figure~\ref{fig:per_class_f1} complements the aggregate metrics by showing
where the strongest main-benchmark move model still struggles. The lifted
ST-GCN with no temporal jitter maintains strong performance on common
footwork and terminal actions, but the long tail remains difficult: fl\`eche
and several blade-line categories still show high variance across folds. This
matters for decision support because these rare events often have
disproportionate semantic impact downstream. The perception interface
therefore benefits not only from higher average F1, but from clearer
visibility into which event types remain brittle.

\subsection{Framework implications}

For this case study, the important outcome is not only which backbone is best,
but that the front end exposes usable trade-offs. Lifted ST-GCN with no
temporal jitter is the strongest move-sensitive default. PoseC3D-style offers
the best blade-line/ECE trade-off among the selected 3D models, while lifted
FERA-CLS no-jitter is the strongest in-house transformer and its no-sampler
variant remains the more calibration-focused transformer choice. The corrected
RGB-fusion and hybrid models still did not justify their added complexity on
the available data. Taken together, these results support treating perception
as a replaceable front end chosen for the needs of the larger decision-support
system.

\subsection{Structured end-to-end decision baseline}
\label{sec:structured_e2e}

Alongside the perception benchmark, we evaluate a compact structured
classifier on top of the fixed 2D ST-GCN token stream from the canonical
no-jitter operating point. Although the lifted 3D branch improves trimmed
perception, additional untrimmed reruns did not improve this downstream
decision stage, so we retain the stronger 2D token stream here. For each of
the 484 exchanges, left/right segment lists are converted into summary features including
per-move counts and confidences, earliest-action indicators, blade-line
summaries, pairwise left/right contrasts, hit values, and an optional
exchange-interaction summary. We compare a multinomial logistic-regression
baseline against random forests with progressively simplified feature-family
subsets under the same grouped match-pool protocol used in the robustness
study.

Table~\ref{tab:structured_e2e} makes three points concrete. First, the random
forest remains the strongest lightweight model family on the pruned 2D token
summary, outperforming both multinomial logistic
regression (0.541 accuracy, 0.512 macro-F1) and HistGradientBoosting (0.599
accuracy, 0.561 macro-F1). Second, the RF feature-family ablation is
substantive. The full-feature RF reaches 0.614
accuracy and 0.614 macro-F1; restricting the model to the three strongest
families --- segment counts, move confidences, and hit values --- already
improves the result to 0.622 accuracy and 0.626 macro-F1; removing blade
summaries gives the highest raw accuracy at 0.626 with slightly lower
macro-F1; and removing the interaction-summary block gives the best balanced
operating point at 0.624 accuracy and 0.632 macro-F1. We therefore adopt this
pruned random forest as the paper's
structured baseline.

Third, the direct 2D-vs-3D token-stream comparison explains why the structured
layer is intentionally kept on the 2D stream even though the lifted branch is
the better trimmed perception model. Replacing only the upstream token stream
with the lifted 3D ST-GCN cache leaves overall accuracy nearly unchanged
(0.624 to 0.622) but reduces macro-F1 from 0.632 to 0.557 and None F1 from
0.650 to 0.412. In practice, the 2D structured baseline recovers 13 of 25
None exchanges, whereas the 3D-token version recovers only 7. This suggests
that the lifted branch sharpens move recognition and even slightly improves
boundary alignment, but the resulting exchange summaries are less
stable for this simple fixed-length downstream classifier. Macro-F1 remains
important here because the output is a 3-class task: it penalizes systems
that perform well on the dominant Left/Right outcomes but collapse on the
rarer None class. The resulting performance is still not high enough to frame
FERA as a competitive autonomous referee, but it does show that the
perception interface supports a nontrivial end-to-end decision layer without
relying on free-form language generation.

\begin{table}[!tb]
  \centering
  \caption{Structured decision results on the final three-way task \{Left, Right, None\} under grouped match-pool evaluation. Top: lightweight model families on the pruned 2D token summary. Middle: RF feature-family ablations. Bottom: the same pruned RF with the lifted 3D ST-GCN token stream.}
  \label{tab:structured_e2e}
  \scriptsize
  \setlength{\tabcolsep}{3pt}
  \resizebox{\columnwidth}{!}{%
  \begin{tabular}{lccc}
    \toprule
    Model & Accuracy & Macro-F1 & None F1 \\
    \midrule
    Logistic regression (2D, pruned) & 0.541 & 0.512 & 0.441 \\
    HistGradientBoosting (2D, pruned) & 0.599 & 0.561 & 0.471 \\
    Random forest (2D, -interaction summary) & 0.624 & \textbf{0.632} & \textbf{0.650} \\
    \midrule
    RF (2D, all families) & 0.614 & 0.614 & 0.615 \\
    RF (2D, counts+move+hit) & 0.622 & 0.626 & 0.634 \\
    RF (2D, -blade summaries) & \textbf{0.626} & 0.628 & 0.634 \\
    \midrule
    RF (3D, pruned) & 0.622 & 0.557 & 0.412 \\
    \bottomrule
  \end{tabular}}
\end{table}

\subsection{Structured decision diagnostics}

The decision baseline becomes more informative when inspected as a token
consumer rather than as a single scalar score. The pruned random
forest is nearly balanced between the two dominant scoring sides (Left F1
0.622, Right F1 0.624), while the rarer None class improves to 0.650 F1 but
still remains recall-limited. In practice, this means the system is cautious
when it predicts None, but it still misses many abstention-worthy exchanges.

\begin{figure}[!tb]
  \centering
  \IfFileExists{structured_decision_diagnostics.pdf}{\includegraphics[width=\columnwidth]{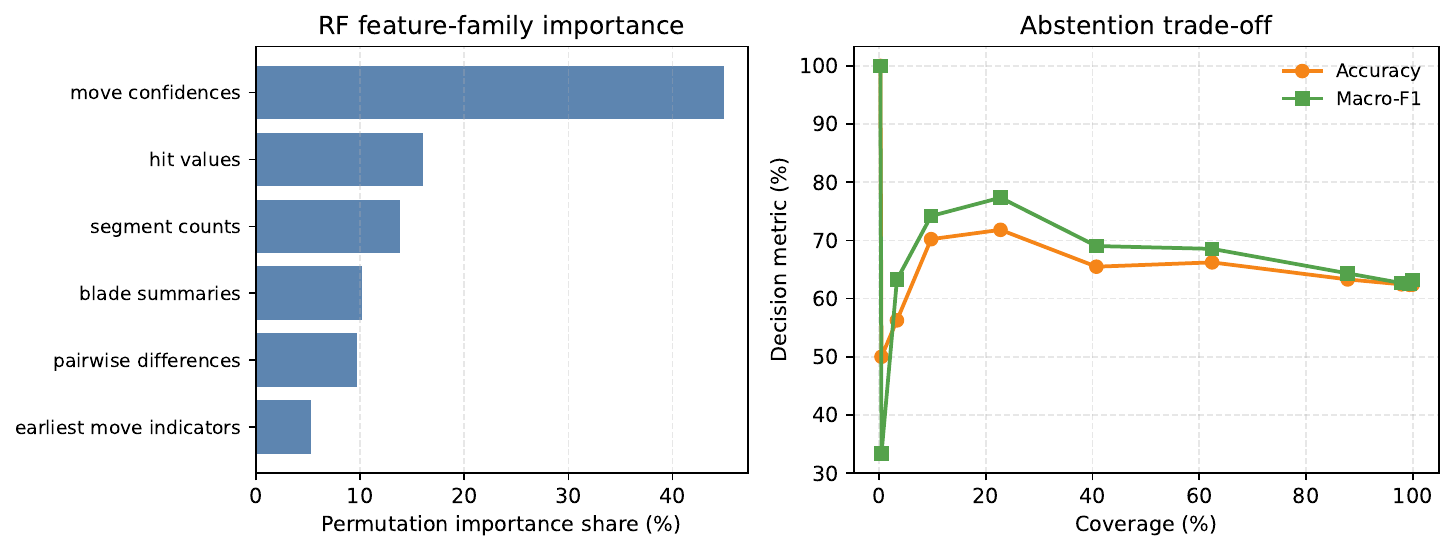}}{\fbox{\parbox[c][0.18\textheight][c]{0.95\columnwidth}{\centering Missing file: \texttt{structured_decision_diagnostics.pdf}}}}
  \caption{Structured-decision diagnostics for the pruned random forest on the fixed 2D ST-GCN token stream. Left: positive permutation-importance share across the remaining feature families after removing the interaction-summary block. Right: abstention trade-off as the minimum retained class confidence increases.}
  \label{fig:structured_diagnostics}
\end{figure}

Figure~\ref{fig:structured_diagnostics} summarizes two additional diagnostics.
First, the permutation-importance analysis and the ablation agree on the same
story: move confidences remain the dominant family, followed by hit values and
segment counts, while earliest-action indicators, pairwise contrasts,
blade-summary features, and the handcrafted interaction block contribute
little or negatively under this fixed protocol. Second, a simple abstention
sweep still reveals a useful caution mode even without formal conformal
prediction. Requiring a maximum class probability above 0.55 keeps 62.4\%
coverage and raises decision accuracy to 0.662, while a stricter 0.65
threshold keeps 22.7\% coverage and raises accuracy to 0.718 with 0.773
macro-F1. This is valuable for practical deployment because it suggests that a
referee-support tool could surface confident exchanges automatically while
deferring ambiguous ones for human review.

\IfFileExists{threshold_generalization_table.tex}{
\subsection{Threshold transfer and boundary quality}

The threshold-transfer check in
Table~\ref{tab:threshold_generalization} addresses a specific reproducibility
concern: are the per-fold per-class thresholds merely overfit knobs, or do they
meaningfully stabilize the deployment-time operating point? The fold-specific
validation thresholds remain the best balanced choice.
Keeping them yields the main structured result (0.624 accuracy, 0.632
macro-F1, None F1 0.650). Replacing them with one shared mean threshold
vector averaged across the ensemble reduces performance substantially to 0.587
accuracy and 0.584 macro-F1. Fixing every move threshold at 0.5 raises raw
accuracy slightly to 0.634, but it also lowers macro-F1 to 0.610, drops None
F1 to 0.556, and increases the no-consensus advance rate from 7.0\% to
12.0\%. In other words, validation tuning is not a cosmetic gain; it mainly
preserves class balance and abstention behavior.

\begin{table}[!tb]
  \centering
  \caption{Threshold-transfer check for the pruned RF on the fixed 2D ST-GCN token stream. Thresholds are validation-tuned per fold, shared by averaging across the ensemble, or fixed globally at 0.5.}
  \label{tab:threshold_generalization}
  \scriptsize
  \setlength{\tabcolsep}{3pt}
  \resizebox{\columnwidth}{!}{%
  \begin{tabular}{lccccc}
    \toprule
    Threshold source & Accuracy & Macro-F1 & None F1 & Seg./exchange & No-consensus \\
    \midrule
    Validation-tuned & 0.624 & \textbf{0.632} & \textbf{0.650} & 20.27 & 0.070 \\
    Fixed 0.5 & \textbf{0.634} & 0.610 & 0.556 & 18.95 & 0.120 \\
    Shared mean threshold & 0.587 & 0.584 & 0.579 & 20.22 & 0.075 \\
    \bottomrule
  \end{tabular}}
\end{table}

We also evaluated the deployment-time segmentation logic directly against the
accepted trimmed segment boundaries. Under the default 2-of-3 dynamic windows,
the 2D pipeline reaches a mean best IoU of 0.381, median best IoU of 0.429,
and F1@0.5 of 0.204, with mean absolute start and end errors of 3.7 and
2.6 frames. The lifted 3D branch is slightly better on this boundary-quality
check (mean best IoU 0.382, median best IoU 0.444, F1@0.5 0.209, mean start
and end errors 3.7 and 2.5 frames), which helps explain why it is the
stronger trimmed perception model. However, the gain is small, and by itself
it does not overcome the downstream 2D-versus-3D token-stream gap in the
structured decision layer. These boundary results therefore make the
deployment defaults more transparent without changing the main conclusion that
the current end-to-end bottleneck is still token quality and None handling
rather than only boundary drift.
}{}

\subsection{Reasoning interface and qualitative examples}

\begin{figure}[!tb]
  \centering
  \subfloat[Left step forward + beat.\label{fig:left_beat}]{%
    \begin{minipage}[t]{0.475\columnwidth}
      \centering
      \IfFileExists{stepforwardbeat.png}{\includegraphics[width=\linewidth]{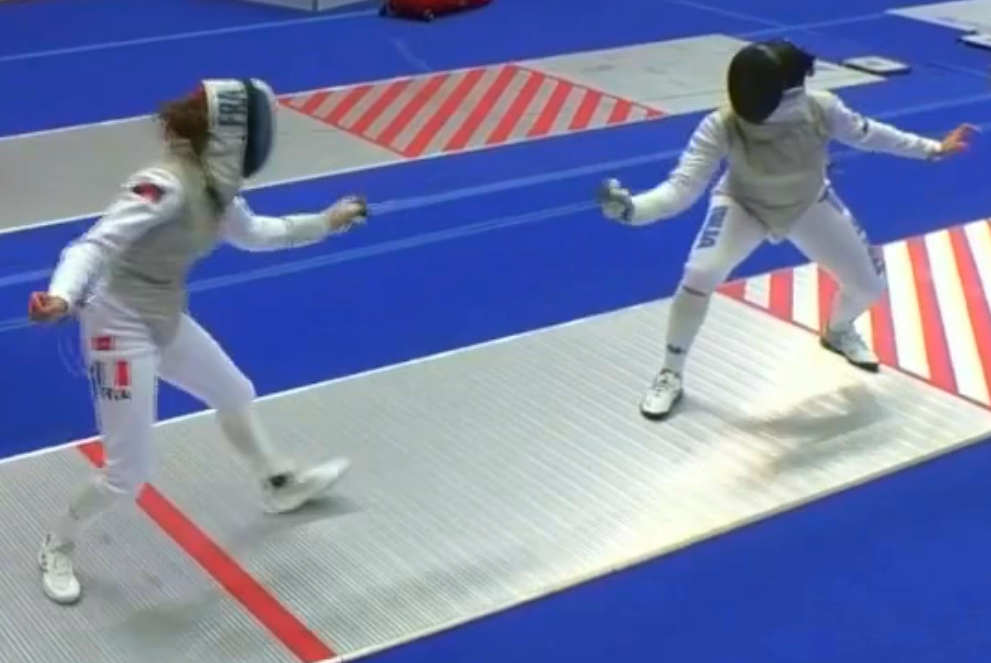}}{\fbox{\parbox[c][0.18\textheight][c]{0.95\columnwidth}{\centering Missing file: \texttt{stepforwardbeat.png}}}}
    \end{minipage}}%
  \hfill
  \subfloat[Left lunge, right counterattack.\label{fig:left_lunge}]{%
    \begin{minipage}[t]{0.475\columnwidth}
      \centering
      \IfFileExists{lunge.png}{\includegraphics[width=\linewidth]{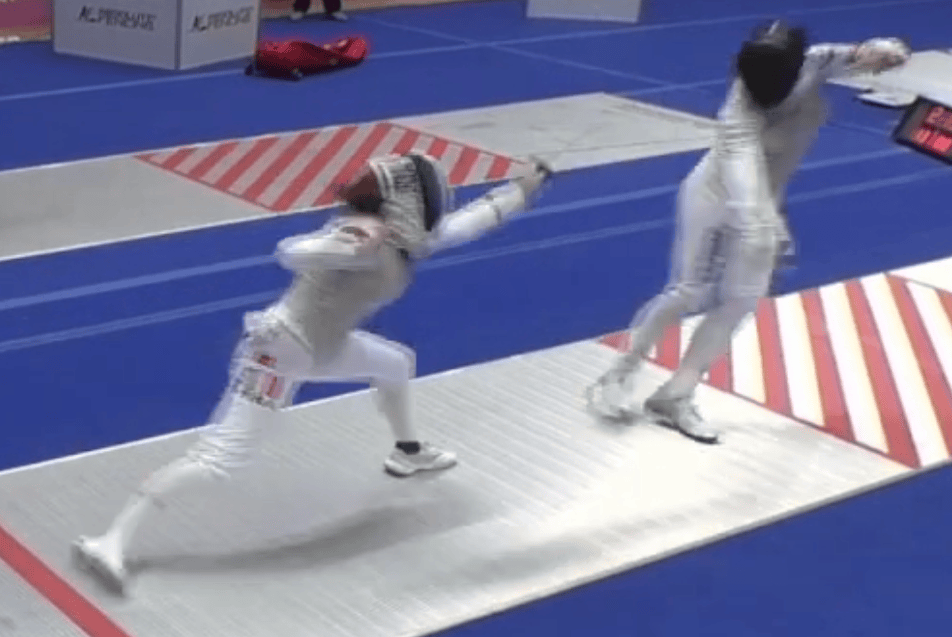}}{\fbox{\parbox[c][0.18\textheight][c]{0.95\columnwidth}{\centering Missing file: \texttt{lunge.png}}}}
    \end{minipage}}
  \caption{Representative frames used to illustrate the perception-to-reasoning interface. The goal here is to show the structured token stream consumed by the downstream module, not to claim state-of-the-art end-to-end officiating accuracy.}
  \label{fig:pipeline_frames}
\end{figure}

Figure~\ref{fig:pipeline_frames} illustrates the intended use of the
perception-to-reasoning pipeline. In both examples, FERA-MBD produces
time-ordered move and blade-line tokens for each fencer. The structured
decision layer consumes those tokens quantitatively, while FERA-LM combines
the same token stream with hit information and retrieved rule text to produce
a human-readable explanation. The figure is therefore best read as an
interface example: it shows what the downstream modules receive once the
perception stage has been fixed.

\section{Discussion}

\subsection{Why the framework extends beyond fencing}

Although developed around foil officiating, FERA is best viewed here as a
case-study pipeline for rule-governed video understanding. The transferable
ideas are straightforward: canonical participant-centric views reduce
left/right nuisance variation, tokenization separates visual evidence from rule
application, calibrated temporal perception keeps uncertainty visible, and a
rule interface lets downstream consumers reason over explicit state estimates
instead of raw frames. This pattern is relevant anywhere decisions depend on
temporally ordered body states and explicit policies, including other
rule-governed sports, structured rehabilitation feedback, instructional skill
assessment, and procedure monitoring.

The experiments also suggest why the perception layer behaves as it does.
Canonical left-oriented views simplify the input distribution; graph-temporal
models match the short coordinated motion patterns common in foil; and a small
depth sidecar adds useful disambiguating signal without redesigning the entire
pipeline. By contrast, the corrected RGB-fusion comparison shows that a simple
late-fusion appearance branch was still not enough to recover blade-sensitive
cues reliably under the present data scale and broadcast-video quality. The
broadened graph comparison also makes one thing clearer: the lifted ST-GCN
result is not merely winning against weak comparators. Even under the same
24D sidecar, the additional 2s-AGCN-style, MS-G3D-style, ResGCN-style, and
PartAtt-style baselines remained well below it. The more reliable operating
point therefore remained the pose-centric front-end, with richer weapon-aware
visual modeling left for future work.

\subsection{Design trade-offs for deployment}

The main deployment trade-off is therefore not a single scalar score but an
operating point. Lifted ST-GCN is the strongest choice when move sensitivity
and overall combined recognition matter most. PoseC3D-style is attractive when
blade-line coverage and ECE matter more. Lifted FERA-CLS no-jitter is the best
in-house transformer for recognition, while its no-sampler variant prioritizes
calibration. Larger multimodal and hybrid models are not automatically
preferable in this data regime, because the dataset is still modest and the
broadcast videos provide limited reliable blade appearance. The front-end
should therefore be selected by downstream needs rather than by architectural
novelty alone.

\subsection{Limitations}

Our study has several limitations that point toward promising directions for
future work. The strongest evidence in this paper remains on the perception
interface. Even after the feature-family ablation, the structured layer reaches
only 0.624 accuracy and 0.632 macro-F1, and FERA-LM is retained primarily as
an explanation-oriented downstream consumer rather than as the main decision
engine. Likewise, although the accepted subset was fully reviewed by a second
annotator and adjudicated jointly, this still falls short of a fully
independent dual-annotation study with blinded relabeling and agreement
statistics collected before discussion.

First, data scale and imbalance remain significant. Manual
frame-level labeling limits the number of segments, and rare actions (e.g.,
fl\`eche and uncommon blade positions) are underrepresented. We expect further
gains with additional data, particularly for long-tail actions.

Second, the annotation scheme deliberately intervalizes co-occurring actions
with shared boundaries. This makes the labeling problem manageable and matches
how the temporal detector is trained, but it also compresses real temporal
offsets between overlapping events such as beat, parry, and locomotion. A
future dataset with independently timed action layers could test whether finer
boundary supervision improves downstream reasoning.

Third, FERA still relies on generic monocular pose estimates and does not
capture the blade explicitly. The main benchmark now includes a lightweight
24D z-only lift sidecar, and that addition clearly improves the best graph
model, but it still provides only coarse relative-depth cues rather than
domain-adapted 3D reconstruction, explicit weapon geometry, or contact.
Fencing-specific detectors, stronger 3D lifting, alternative lifters, or
explicit hand/weapon keypoints may still improve robustness to viewpoint and
occlusion and provide more accurate information about blade orientation and
contact.

Fourth, the reasoning layer is only as strong as the token stream it receives.
When upstream temporal tokens omit decisive events or encode them too
coarsely, downstream decisions remain brittle. Improving this stage likely
requires richer event-state supervision, more explicit decision annotations,
and better temporal grounding, not merely larger language models. Collecting
referee justifications together with event-level labels would also enable a
more direct evaluation of explanation quality and final decision consistency.
For this submission, we therefore treat the structured decision layer as the
quantitative end-to-end consumer and FERA-LM as explanation-only; we do not
yet report a quantitative benchmark for explanation quality or rule retrieval.

Finally, we still do not have explicit athlete identities for every clip, so
the grouped robustness test remains a match-disjoint proxy rather than a
strict athlete-disjoint benchmark. It is stronger than pure clip-level
splitting, and the threshold-transfer and boundary-quality checks make the
deployment-time defaults less opaque, but none of these analyses substitutes
for a true identity-disjoint or cross-competition transfer study. The
available filename prefixes are too coarse and too uneven to serve as a
credible substitute for athlete- or competition-disjoint evaluation.

\section{Conclusion}

FERA presents a pose-based pipeline for multimedia decision support in
rule-governed settings, studied here through foil fencing. The central idea is
to expose a stable boundary between uncertain perception and downstream
reasoning: canonicalize each participant, convert motion into compact tokens,
infer temporally localized action/state labels with calibrated confidence, and
reason over those tokens rather than over raw video.

Under a fixed protocol, a lightweight 24D z-only depth sidecar materially
strengthens the perception front-end. Lifted ST-GCN with no temporal jitter is
the strongest overall operating point, and the added 2s-AGCN-style,
MS-G3D-style, ResGCN-style, and PartAtt-style baselines do not displace it.
PoseC3D-style and the lifted FERA-CLS variants define alternative blade- and
calibration-oriented trade-offs. On top of the fixed 2D ST-GCN token stream, a
compact pruned random-forest decision layer improves the final decision
benchmark to 0.624 accuracy and 0.632 macro-F1 without relying on language
generation; direct 2D-vs-3D decision comparisons show that the stronger 3D
perception branch does not automatically yield a stronger downstream token
stream for this simple classifier.

The broader contribution is therefore architectural rather than sport-specific:
a reproducible case study of how to connect pose-derived evidence, uncertainty
estimates, structured decision logic, and explanation-oriented rule grounding.
Future work should expand the event vocabulary, strengthen weapon-specific
sensing, collect richer decision-level supervision, and test how well the same
interface transfers to other domains where video understanding must support
explicit decisions rather than only label isolated actions.

\bibliographystyle{IEEEtran}
\bibliography{main}

\end{document}